\PassOptionsToPackage{table}{xcolor}

\documentclass{Interspeech2024}
\usepackage{amsmath,graphicx}
\usepackage{adjustbox}
\usepackage{url}
\usepackage{array}
\usepackage{multirow}
\usepackage{amssymb}
\usepackage{pifont}
\usepackage{pgfplots}
\usepackage{float} % here for H placement parameter
\usepackage{multirow}
\usepackage{booktabs}
\usepackage{makecell}
\usepackage{tikz}
\usepackage{tipa}
\usepackage{setspace}
\usepackage{cite}
\usepackage{xcolor}
\usepackage{caption}
\usepackage{subcaption}
\usepackage{lipsum}

\interspeechcameraready

\title{SyncVSR: Data-Efficient Visual Speech Recognition with \\ End-to-End Crossmodal Audio Token Synchronization}

\name{Young Jin Ahn$^{1\star}$, Jungwoo Park$^{2\star}$, Sangha Park$^3$, Jonghyun Choi$^4$, Kee-Eung Kim$^1$}

\address{
  $^1$KAIST
  $^2$Kwangwoon University
  $^3$Ajou University
  $^4$Seoul National University
}

\email{snoop2head@kaist.ac.kr, 
affjljoo3581@kw.ac.kr, 
wkrtkd222@ajou.ac.kr, 
jonghyunchoi@snu.ac.kr,
kekim@kaist.ac.kr
}

\keywords{visual speech recognition, lip-reading, crossmodal learning, end-to-end training, data efficiency}

\newcommand{\cmark}{\color{green} \ding{51}}%
\newcommand{\xmark}{\color{red} \ding{55}}%

\pgfplotsset{width=3.9cm,compat=1.9}
\definecolor{forestgreen}{RGB}{22,139,22}
\newcommand{\etal}{\textit{et al}.}
\definecolor{lightblue}{rgb}{0.93,0.95,1.0} % Define custom color
\newcommand\blfootnote[1]{%
  \begingroup
  \renewcommand\thefootnote{}\footnote{#1}%
  \addtocounter{footnote}{-1}%
  \endgroup
}

\begin{document}

\maketitle


\begin{abstract}
Visual Speech Recognition (VSR) stands at the intersection of computer vision and speech recognition, aiming to interpret spoken content from visual cues. A prominent challenge in VSR is the presence of homophenes—visually similar lip gestures that represent different phonemes. Prior approaches have sought to distinguish fine-grained visemes by aligning visual and auditory semantics, but often fell short of full synchronization. To address this, we present SyncVSR, an end-to-end learning framework that leverages quantized audio for frame-level crossmodal supervision. By integrating a projection layer that synchronizes visual representation with acoustic data, our encoder learns to generate discrete audio tokens from a video sequence in a non-autoregressive manner. SyncVSR shows versatility across tasks, languages, and modalities at the cost of a forward pass. Our empirical evaluations show that it not only achieves state-of-the-art results but also reduces data usage by up to ninefold.
\end{abstract}

\section{Introduction}
\label{sec:intro}

Visual Speech Recognition (VSR), also referred to as lip-reading, constitutes the process of decoding spoken language through the observation of the visual cues, specifically the movements of the lips and facial dynamics. This technology holds critical importance in a variety of contexts, including the interpretation of lip movements from individuals with speech disorders~\cite{Laux2023TwostageVS}, benefiting individuals with hearing disorders~\cite{TyeMurray2006AudiovisualIA}, recognizing spoken content within environments where acoustic signals are compromised~\cite{Martinez2020Lipreading,Xu2020DiscriminativeMS}, providing voiceovers to silent historical films, and fortifying security systems~\cite{Haliassos2020LipsDL}.\blfootnote{$\star$ Equal contribution.}

\begin{figure}
\centering
\centerline{\includegraphics[width=8cm]{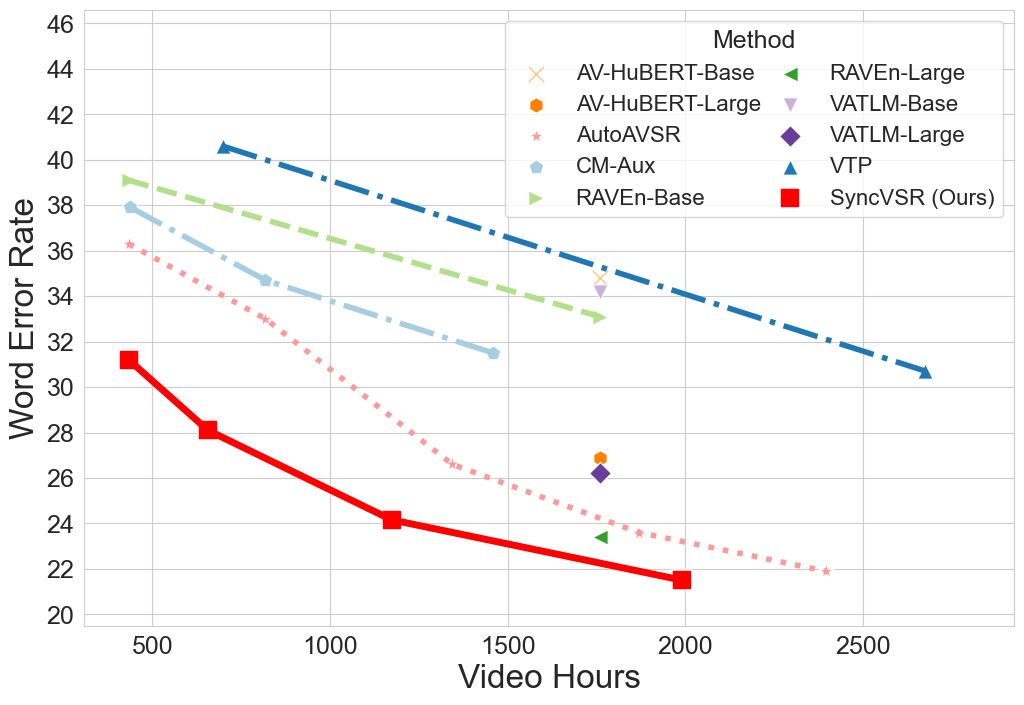}\hspace{0.2cm}}
\vspace*{-10pt}\caption{Performance of SyncVSR on LRS3\cite{afouras2018lrs3ted} benchmark. SyncVSR outperforms available methods given the similar amount of video data resources. Our method also advances a tier in model size, where our base-size model shows superior performance compared to other large-size models.}
\label{fig:LRS3-Data-Scaling}
\end{figure}

The primary challenge encountered in VSR stems from the inherent scarcity of information that can be extracted from visual cues alone~\cite{ren2021learning}. Central to this issue is the presence of homophenes, wherein disparate sounds are visually manifested through identical or nearly identical lip movements~\cite{Kim2022Distinguishing}. Such phenomena represent significant ambiguity in the analysis of visemes, the fundamental units of visual speech recognition. This ambiguity poses considerable difficulties, as it muddles the clarity of speech interpretation through visual means alone. 

Previous research to overcome such limitations has predominantly revolved around aligning visual with auditory semantics, attempting to reduce the gap between audio models and visual models. Earlier techniques~\cite{zhao2020hearing, Afouras2018DeepAS, ma2022visual} harnessed knowledge distillation from pretrained Automatic Speech Recognition (ASR) systems. Subsequent studies~\cite{Kim2022Distinguishing, Yeo2023mtlam, Minsu2021Multi} trained auxiliary audio modules along with the visual encoder in order to transfer speech knowledge. However, the aforementioned methods create indirect links to the acoustic data, as visual models interact with the semantics of audio encoders rather than the acoustic data. Such audio modules might provide insufficient hidden knowledge to the visual modules due to the crossmodal gap in representations~\cite{ren2021learning}. Recent works~\cite{Ma2021LiRA, shi2022avhubert, kim2023lowresource} strived to connect the visual encoder with speech data, but they utilized handcrafted features (e.g., spectrograms or MFCCs) as their inputs or targets, which possibly encompass inductive biases that could affect the learned representations~\cite{haliassos2023raven}.

Moreover, several works~\cite{haliassos2023raven, kim2023lowresource, shi2022avhubert, Zhu2022VATLMVP} introduced learning methods based on crossmodal masked reconstruction, where portions of visual inputs are replaced with masked frames and models are trained to reconstruct corresponding audio representations. Nevertheless, an alternative method to masked segment reconstruction has been introduced in the Natural Language Processing (NLP) domain, which is token-level discrimination~\cite{He2021DeBERTaV3ID, clark2020electra}. A key advantage of such discriminatory supervision is that the model learns from all input tokens instead of just the small masked-out subset, advancing a tier of performance and being more sample-efficient~\cite{clark2020electra}. This method is even more promising in the VSR domain since there is a fine-grained correspondence between the visual and auditory modalities, which provides a natural source of self-supervision~\cite{haliassos2023raven}.

In light of the above, we propose the SyncVSR framework, an innovative approach to VSR that directly aligns visual phonetic units with their acoustic counterparts through quantized audio tokens, facilitating robust end-to-end crossmodal synchronization. By exploiting the discrete nature of quantized audio for frame-level supervision, SyncVSR circumvents the limitations of previous methods that rely on indirect semantic alignment or utilize potentially biased handcrafted features. This allows our model to discern fine-grained phonetic differences inherent in homophenes, enhancing the model's interpretative fidelity and data efficiency. Our empirical results highlight the efficacy of SyncVSR, which establishes new benchmark results across a range of VSR tasks.

\begin{figure}
\centering
\centerline{\hspace{0.25cm}\includegraphics[width=7cm]{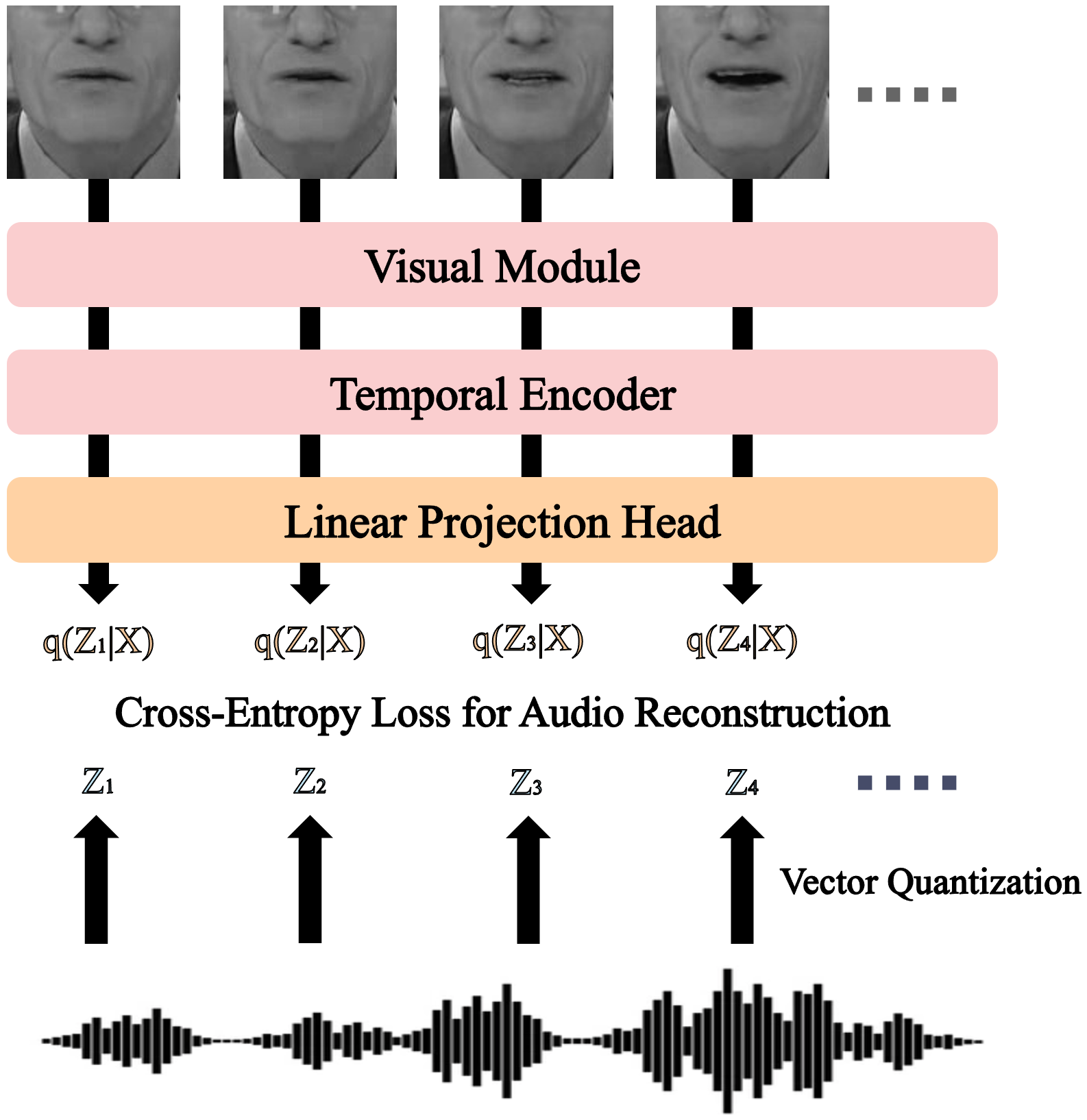}}
\vspace*{-5pt}\caption{Overview of the SyncVSR training framework. Given a sequence of video frames, the encoder generates a corresponding sequence of quantized audio tokens in a non-autoregressive manner. $z_t$ denotes audio tokens, and $q(z_t|x)$ is the encoder's prediction through a linear projection layer.}
\label{fig:SyncVSR}
\end{figure}

\section{Methodology}
\label{sec:SyncVSR}

\noindent\textbf{Crossmodal Audio Token Synchronization.} Our work integrates audio reconstruction loss with VSR training objectives. Conventionally, word-level VSR employs a word classification loss, whereas sentence-level VSR utilizes the joint CTC-Attention loss~\cite{Kim2017JointCTC}. The total loss is the weighted sum of the task-specific objective loss and our audio reconstruction loss.

Let $\mathcal{D}$ be a training set that consists of a training sample $(x, y, z)$. Let $x$ and $y$ be the input video and ground truth label. Let $z=\{ z_t \} _{t\leq T}$ be a discrete audio sequence corresponding to the input video $x$.

\noindent\textbf{Word Classification Loss.} For word-level VSR, cross-entropy loss measures the difference between predicted class probabilities and the ground truth labels. Given that $y$ represents the ground truth category for the input video $x$, the objective loss is formulated as follows:
\begin{equation*}
\mathcal{L}_{\text{task}} = - \mathbb{E}_{(x, y, z) \in \mathcal{D}} \log p(y|x)
\end{equation*}
where $p(y|x)$ denotes the output probability from the model.

\noindent\textbf{Joint CTC-Attention Loss.} For sentence-level VSR, we employ a combination of Connectionist Temporal Classification (CTC)~\cite{graves2006connectionist} loss from the encoder and Language Modeling (LM) loss from the decoder, known as joint CTC-Attention loss.

Let $\pi = \{\pi_t\}_{t \leq T}$ be intermediate CTC labels, and $\phi(y)$ be a set of all possible intermediate labels for CTC loss. Using $p_\text{LM}$ for language modeling and $p_\text{CTC}$ for conditional independent prediction, we define the losses as follows:
\begin{equation*}
    \mathcal{L}_{\text{LM}} = - \mathbb{E}_{(x, y, z) \in \mathcal{D}} \sum_{t \leq T}\log p_\text{LM}(y_t|x, y_{< t}),
\end{equation*}
\begin{equation*}
    \mathcal{L}_{\text{CTC}} = - \mathbb{E}_{(x, y, z) \in \mathcal{D}} \sum_{\pi \in \phi(y)} \sum_{t \leq T}\log p_\text{CTC}(\pi_t|x) ,
\end{equation*}
and the final objective loss is defined as a combination of $\mathcal{L}_\text{LM}$ and $\mathcal{L}_\text{CTC}$, i.e.,
\begin{equation*}
    \mathcal{L}_{\text{task}} = \alpha \mathcal{L}_{\text{CTC}} + (1 - \alpha) \mathcal{L}_{\text{LM}}
\end{equation*}
where $\alpha$ is a hyperparameter with the constraint $0 \leq \alpha \leq 1$.

\begin{figure}
    \centering
    \centerline{\includegraphics[width=7cm]{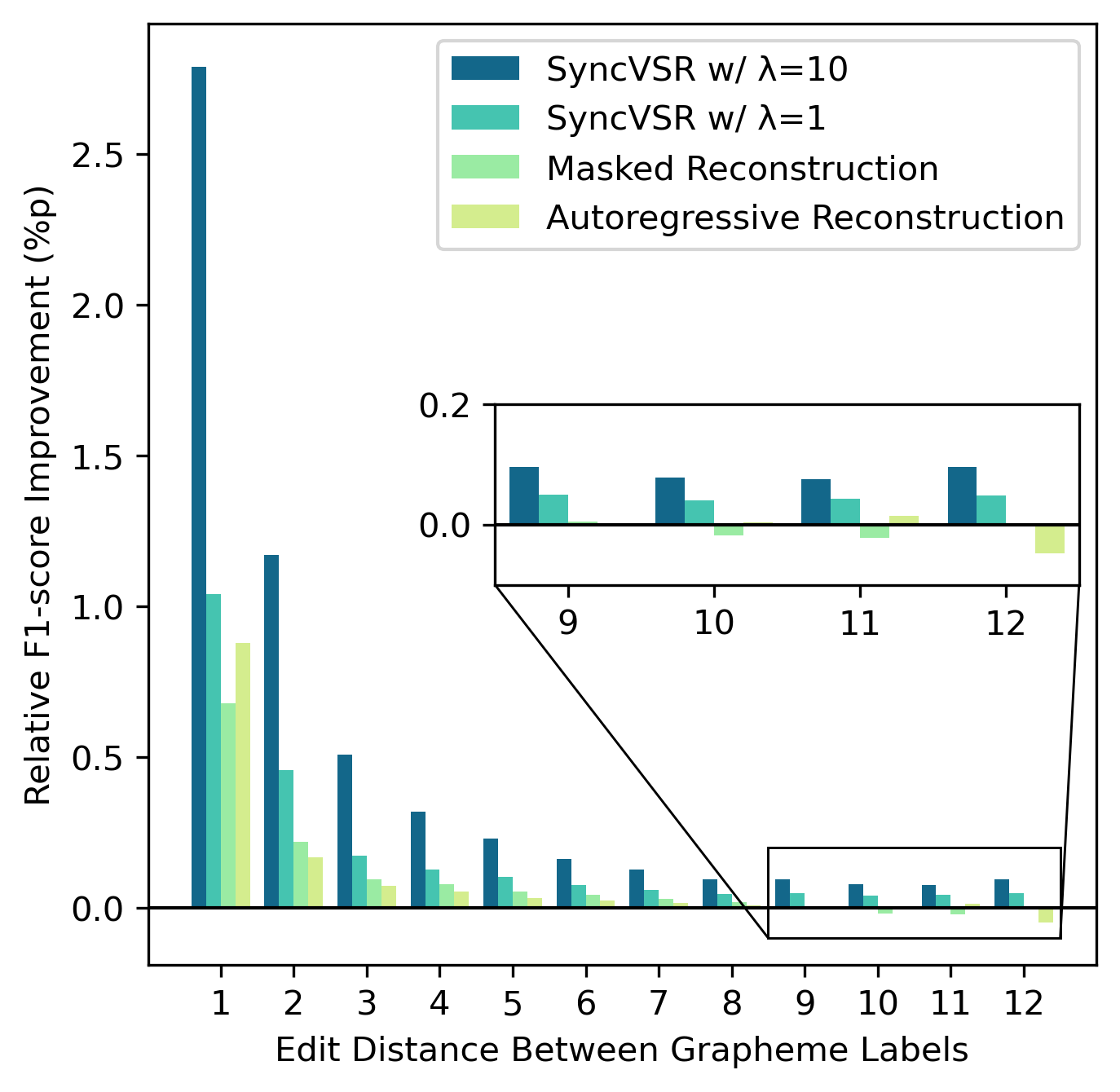}}
    \vspace*{-10pt}\caption{The edit distance of word pairs and the model's discriminative ability. Homophene pairs resemble each other closely in graphemes, a scenario where SyncVSR shows better classification performance over the vanilla setting trained without audio information. Non-autoregressive generation with strong audio reconstruction loss weight ($\lambda$) is optimal, whereas masked reconstruction could cause harm in certain instances.}
    \label{fig:edit_distance}
\end{figure}

\noindent\textbf{Audio Reconstruction Loss.} 
The synchronization between audio and video in our framework is designed to align each video frame with a corresponding number of audio tokens. This alignment is based on audio (16kHz) and video (25fps) sampling rates, with a specific hop size ensuring a coherent match between video frame rate and audio units. We use a ratio of one video frame to four vector quantized audio tokens (100Hz).

To make the model generate discrete audio tokens, we use cross-entropy loss to predict the quantized audio $z_t$ from the input video frames. Let $q(z_t|x)$ be an output of the model at time $t$ from the video $x$. The audio reconstruction loss is as follows:
\begin{equation*}
\mathcal{L}_{\text{Sync}} =\mathbb{E}_{(x, y, z) \in \mathcal{D}} \left[ -\frac{1}{T}\sum_{t \leq T} \log q(z_t|x) \right]
\end{equation*}
where $T$ denotes the number of video frames.

Using these, we simply design the total loss as below, with the hyperparameter $\lambda$ as the weight for the audio reconstruction loss,
$$\mathcal{L}=\mathcal{L}_{\text{task}}+\lambda\mathcal{L}_{\text{Sync}}.$$ 

\begin{table*}[ht!]
\caption{Video-based VSR performance on word-level tasks. Evaluations were done on Lip Reading in the Wild (LRW)~\cite{chung2017LRW} English benchmark and CAS-VSR-W1K~\cite{Yang2018LRW1000AN} Chinese benchmark. WB implies the usage of word boundary, which is an indicator for the target word's appearance. Our metrics are averaged across three experiments, with subscripts notating the standard deviation. Transcription Alignment is our experiment using the method of aligning character-level pseudo-labels from ASR models instead of auditory data.}
  \label{tab:word_level}
  \centering
  \begin{tabular}{lccccc}
    \toprule
    \multirow{2.5}{*}{\textbf{Method}} & \multirow{2.5}{*}{\textbf{Temporal Model}} & \multirow{2.5}{*}{\textbf{Video Hours}} & \multicolumn{3}{c}{\textbf{Top-1 Acc. (\%) $\uparrow$}} \\
    \cmidrule(lr){4-6}
    & & & \textbf{LRW} & \textbf{LRW(WB)} & \textbf{CAS-VSR-W1K} \\
    \midrule
Born-Again~\cite{Ma2020TowardsPL} & MS-TCN & 157 & 87.9 & - & 46.6 \\
LiRA~\cite{Ma2021LiRA} & Conformer & 590 & 88.1 & - & - \\
WPCL + APFF~\cite{Tian2022LipreadingMB} & MS-TCN & 157 & 88.3 & - & - \\
Ma \etal~\cite{Ma2020LipreadingWD} & DC-TCN & 157 & 88.4 & - & 43.7 \\
Feng \etal~\cite{feng2020learn} & BiGRU & 157 & 86.2 & 88.4 & \underline{55.7} \\
MVM~\cite{Kim2022Distinguishing} & MS-TCN & 157 & 88.5 & - & 53.8 \\
NetVLAD~\cite{Yang2022ImprovedWL} & MS-TCN & 157 & 89.4 & - & - \\
Koumparoulis \etal~\cite{Koumparoulis2022Accurate} & Transformer & 157 & 89.5 & - & - \\
Training Strategy~\cite{Ma2022Training} & DC-TCN & 157 & 90.4 & 92.1 & - \\
MTLAM~\cite{Yeo2023mtlam} & DC-TCN & 157 & \underline{91.7} & - & 54.3 \\
Training Strategy + LiRA & DC-TCN & 595 & - & 92.3 & - \\
Training Strategy + CM-Aux~\cite{ma2022visual} & DC-TCN & 1,459 & - & \underline{92.9} & - \\

\rowcolor{lightblue}Transcription Alignment & Transformer & 157 & 93.1 & 94.8& - \\
\rowcolor{lightblue}\textbf{SyncVSR} & Transformer & 157 & \textbf{93.2}~\scriptsize{$\pm$ 0.1} & \textbf{95.0}~\scriptsize{$\pm$ 0.0} & \textbf{58.2}~\scriptsize{$\pm$ 0.0} \\

    \bottomrule
  \end{tabular}
\end{table*}

\begin{table}[th]
\caption{Landmark-based VSR evaluated on a word-level task. The methods below were trained from scratch on the LRW.}
  \label{tab:landmark_word_level}
  \centering
  \resizebox{\columnwidth}{!}{%
  \begin{tabular}{lcccc}
    \toprule
    \textbf{Method} & \textbf{Input Type} & \textbf{\#Params} & \textbf{Top-1 $\uparrow$} \\
    \midrule

Lip Graph Assisted~\cite{Liu2020LipGA} & Graph & 30M & 49.3 \\
Adaptive GCN~\cite{Sheng2022AdaptiveSG} & Graph & 45M & 60.7 \\
Another Point of View~\cite{Pouthier2023AnotherPO} & Pointcloud & 12M & \underline{62.7} \\
\rowcolor{lightblue}\textbf{SyncVSR} & Pointcloud & 11M & \textbf{75.1}~\scriptsize{$\pm$ 0.1} \\
\rowcolor{lightblue}\textbf{SyncVSR(WB)} & Pointcloud & 11M & \textbf{80.3}~\scriptsize{$\pm$ 0.0} \\

    \bottomrule
  \end{tabular}}
\end{table}

\begin{table}[ht!]
\caption{Video-based VSR performance on sentence-level tasks grouped with video data resource usage. Evaluations were done on LRS2~\cite{LRS2} and LRS3~\cite{afouras2018lrs3ted} benchmarks. LM indicates whether an external language model is used. The methods listed below use the base-size model unless specified as large-size. Reported scores have a standard deviation smaller than 0.5.}
  \label{tab:sentence_level}
  \centering
  \resizebox{\columnwidth}{!}{%
  \begin{tabular}{lcccc}
    \toprule
    \multirow{2.5}{*}{\textbf{Method}} & \multirow{2.5}{*}{\textbf{Video Hours}} & \multirow{2.5}{*}{\textbf{LM}} & \multicolumn{2}{c}{\textbf{WER $\downarrow$}} \\
    \cmidrule(lr){4-5}
    & & & \textbf{LRS2} & \textbf{LRS3} \\
    \midrule
    \rowcolor{gray!20}\multicolumn{5}{c}{\textbf{Less than 500h}} \\
TDNN~\cite{Yu2020AudioVisualRO} & 223 & \cmark & 48.9 & -\\
CM-Seq2Seq~\cite{Ma2021EndConformer} & 223/438 & \cmark & 39.1 & 46.9 \\
CM-Aux~\cite{ma2022visual} & 223/438 & \cmark & \underline{32.9} & 37.9 \\
RAVEn~\cite{haliassos2023raven} & 438 & \cmark & -  & 39.1 \\
AutoAVSR~\cite{ma2023autoavsr} & 438 & \cmark & -  & \underline{36.3} \\
\rowcolor{lightblue}\textbf{SyncVSR (Ours)} & 223/438 & \xmark & \textbf{30.7} & \textbf{33.3} \\
\rowcolor{lightblue}\textbf{SyncVSR (Ours)} & 223/438 & \cmark & \textbf{28.9} & \textbf{31.2} \\
\midrule
    \rowcolor{gray!20}\multicolumn{5}{c}{\textbf{Less than 1000h}} \\
KD + CTC~\cite{Afouras2019ASRIA} & 995 & \cmark & 51.3 & 59.8 \\
KD-Seq2Seq~\cite{ren2021learning} & 818 & \xmark & 49.2 & 59.0 \\
MVM~\cite{Kim2022Distinguishing} & 818 & \xmark & 44.5 & - \\
LiRA~\cite{Ma2021LiRA} & 661 & \cmark & 38.8 & - \\
RAVEn & 661 & \cmark & 32.1 & - \\
VTP~\cite{prajwal2022sub} & 698 & \cmark & 28.9 & 40.6 \\
AutoAVSR~\cite{ma2023autoavsr} & 818 & \cmark & 27.9 & \underline{33.0} \\
CM-Aux & 818 & \cmark & \underline{27.3} & 34.7 \\
\rowcolor{lightblue}\textbf{SyncVSR (Ours)} & 661 & \xmark & \textbf{22.0} & \textbf{30.4} \\
\rowcolor{lightblue}\textbf{SyncVSR (Ours)} & 661 & \cmark & \textbf{20.0} & \textbf{28.1} \\
\midrule
    \rowcolor{gray!20}\multicolumn{5}{c}{\textbf{Less than 2000h}} \\
TM-Seq2Seq~\cite{Afouras2018DeepAS} & 1,391 & \cmark & 48.3 & 58.9 \\
CM-Aux & 1,459 & \cmark & 25.5 & 31.5 \\
AV-HuBERT~\cite{shi2022avhubert} & 1,992/1,759 & \xmark & 31.2 & 34.8 \\
AV-HuBERT-Large & 1,992/1,759 & \xmark & 25.5 & 26.9 \\
VATLM~\cite{Zhu2022VATLMVP} & 1,992/1,759 & \xmark & 30.6 & 34.2 \\
VATLM-Large & 1,992/1,759 & \xmark & 24.3 & 26.2 \\
LMDecoder\cite{kim2023lowresource} & 1,992 & \xmark & 23.8 & - \\
RAVEn & 1,992/1,759 & \xmark & - & 33.1 \\
RAVEn-Large & 1,992/1,759 & \xmark & 19.3 & 24.4 \\
RAVEn-Large & 1,992/1,759 & \cmark & \underline{17.9} & \underline{23.1} \\
\rowcolor{lightblue}\textbf{SyncVSR (Ours)} & 1,992 & \xmark & 18.5 & 23.4 \\
\rowcolor{lightblue}\textbf{SyncVSR (Ours)} & 1,992 & \cmark & \textbf{16.5} & \textbf{21.5} \\
% \midrule
\midrule
    \rowcolor{gray!20}\multicolumn{5}{c}{\textbf{Greater than 2000h}} \\
VTP & 2,676 & \cmark & 22.6 & 30.7 \\
AutoAVSR & 3,448 & \cmark & \textbf{14.6} & 19.1 \\
ViT 3D~\cite{Serdyuk2022TransformerBasedVF} & 90,000 & \xmark & - & 17.0 \\
LP Conformer~\cite{Chang2023ConformersAA} & 100,000 & \xmark & - & \textbf{12.8} \\
    \bottomrule
  \end{tabular}}
\end{table}

% We didn't use engineering tricks!
\section{Experimental Setup}
\label{sec:experimentalsetup}

\noindent\textbf{Training Dataset.} We employ the LRW~\cite{chung2017LRW} dataset for English and the CAS-VSR-W1K~\cite{Yang2018LRW1000AN} for Chinese to evaluate word-level VSR tasks. The LRW dataset comprises 500 words, each represented by up to 1,000 training videos. The LRW-1000 dataset consists of 718,018 videos spanning 1,000 words. Our sentence-level experimental framework was anchored on the LRS2~\cite{LRS2} and LRS3~\cite{afouras2018lrs3ted} datasets, representing the most extensive publicly available resources for audio-visual speech recognition in English. The LRS2 dataset, sourced from BBC programs, comprises 144,482 video clips, totaling 225 hours of video content. The LRS3 dataset, harvested from TED talks, encompasses 151,819 video clips, amassing 439 hours of footage. Additional training data was sourced from the English-speaking segments of the VoxCeleb2~\cite{Chung2018VoxCeleb2DS} dataset, comprised of a training corpus totaling 1,323 video hours, complemented by transcriptions following the scheme of AutoAVSR~\cite{ma2023autoavsr}.

\noindent\textbf{Dataset Preprocessing.} We used MediaPipe~\cite{lugaresi2019mediapipe} to identify the region of interest with a size of 128 x 128 for video-based VSR, and the extracted landmark data served as input for a pointcloud-based VSR system. We used a data augmentation scheme of a resized random crop with a size of (96, 96) and a random horizontal flip and applied a center crop for inference similar to that of previous works~\cite{Martinez2020Lipreading, ma2023autoavsr}.

\noindent\textbf{Model Architecture.} For word-level VSR, an encoder is composed of a combination of 3D CNN, ResNet18, and Transformer~\cite{vaswani2017attention} to extract video features following the previous works~\cite{Kim2022Distinguishing, Minsu2021Multi, Yeo2023mtlam, Ma2022Training}. On the other hand, Conformer~\cite{gulati20_conformer} is used as a temporal backbone for sentence-level VSR, where we follow the model size and configuration of previous works' settings~\cite{Ma2021LiRA, Ma2021EndConformer, haliassos2023raven}.

\noindent\textbf{Training Recipe.} For word-level VSR tasks, we train the model for 200 epochs with the Adam~\cite{Kingma2014AdamAM} optimizer. The learning rate increased from 0 to 0.0001 for the first 5 epochs and then decreased linearly. In the case of sentence-level VSR, the model is trained for 100 epochs with the Adam optimizer, where the learning rate linearly decays from the peak of 0.001 at the 3rd epoch. Batch size is 384 for word-level and 64 for sentence-level, distributed to 4$\times$A100 GPUs. The rest of the training specifics follow the previous work's settings from~\cite{ma2023autoavsr}. Our metrics were obtained from the average of three random seeds.

\section{Results}

\noindent\textbf{Versatility Across Tasks, Languages, and Modalities.} Our framework is comprehensively evaluated according to tasks, languages, and input modalities. In word-level tasks, shown in Table \ref{tab:word_level}, SyncVSR marks state-of-the-art results in English and Chinese benchmarks. In sentence-level tasks, displayed in Figure \ref{fig:LRS3-Data-Scaling} and Table \ref{tab:sentence_level}, SyncVSR outperforms available methodologies when given a similar amount of video dataset. Notably, our method also advances a tier in model size, where our base-size model shows superior performance over other large-size models. Our method also achieves state-of-the-art performance in landmark-based VSR tasks shown in Table \ref{tab:landmark_word_level} and Table \ref{tab:ablation}.

\noindent\textbf{Distinguishing Homophenes.} Homophenes often closely resemble each other in their graphemes—the smallest functional units of a writing system. For example, homophene pairs, like (\textit{Million}, \textit{Billion}) or (\textit{Living}, \textit{Giving}), differ by just one grapheme. Although earlier research, notably by Kim \etal~\cite{Kim2022Distinguishing} has examined a subset of these pairs, a full-scale evaluation of every potential homophene pair has yet to be achieved. As a result, in Figure \ref{fig:edit_distance}, we assess the relative F1-score gain of existing training methods over a vanilla setting that does not utilize the audio data, focusing on the grapheme edit distances. This suggests that the inclusion of an audio reconstruction loss objective assists in differentiating visemes that are mapped into similar graphemes, which is where homophene pairs are typically found.

\begin{table}[th]
  \caption{Impact of CTC loss and audio reconstruction loss on the LRS2 benchmark. To the best of our knowledge, this is the first instance of reporting a successful implementation of landmark-based sentence-level visual speech recognition.}
  \label{tab:ablation}
  \centering
  \resizebox{\columnwidth}{!}{%
  \begin{tabular}{cccccc}
    \toprule
    \multirow{2.5}{*}{\textbf{Sync}} &\multirow{2.5}{*}{\textbf{CTC}} & \multicolumn{2}{c}{\textbf{WER $\downarrow$}} & \multicolumn{2}{c}{\textbf{Perplexity $\downarrow$}} \\
    \cmidrule(lr){3-4} \cmidrule(lr){5-6}
    & & \textbf{Video} & \textbf{Pointcloud} & \textbf{Video} & \textbf{Pointcloud} \\
    \midrule
    \xmark & \xmark & 45.9 & 99.9 & 4.0 & 40.7 \\ 
    \xmark & \cmark & 43.2 & 99.8 & 4.1 & 40.4 \\ 
    \rowcolor{lightblue}\cmark & \xmark & 38.1 & 77.4 & 2.8 & 8.1\\
    \rowcolor{lightblue}\cmark & \cmark & \textbf{30.7} & \textbf{74.6} & \textbf{2.7} & \textbf{7.7} \\
    \bottomrule
  \end{tabular}}
\end{table}

\begin{figure}
\begin{minipage}[b]{0.48\linewidth}
  \centering
  \centerline{\includegraphics[width=4.15cm]{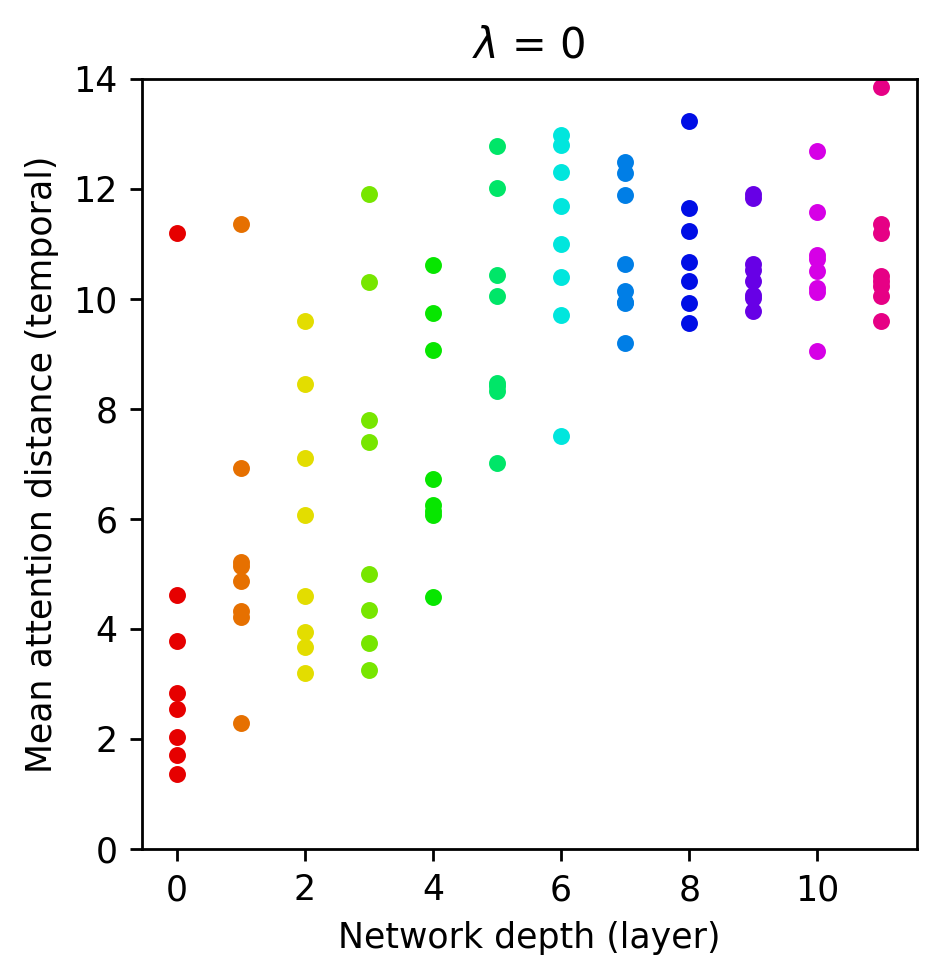}}
  \centerline{(a) Trained w/o Sync}\medskip
\end{minipage}
\hfill
\begin{minipage}[b]{0.48\linewidth}
  \centering
  \centerline{\includegraphics[width=4.15cm]{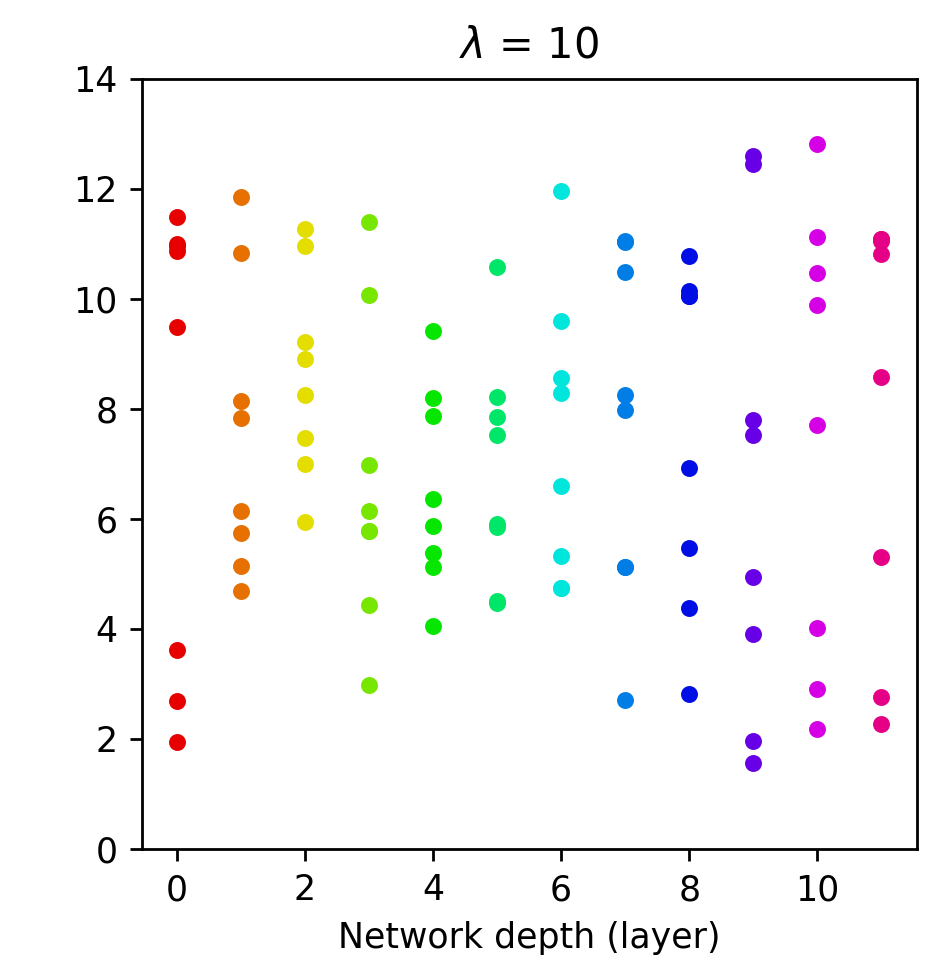}}
  \centerline{(b) Trained w/ Sync}\medskip
\end{minipage}
\vspace*{-12pt}\caption{Influence of audio reconstruction loss weight ($\lambda$) on the encoder's representation visualized with the mean attention distance~\cite{dosovitskiy2021an} distribution. Each point indicates the weighted distance of attention from the query frame to other frames.}
\label{fig:mean_attention_distance}
\end{figure}

\noindent\textbf{Significance of Full Sequence Synchronization.} Furthermore, Figure \ref{fig:edit_distance} compares our reconstruction method with masked reconstruction. Not only does our full-length non-autoregressive generation excel when graphemes are similar, but it is also noteworthy that masked reconstruction could cause harm for classifying pairs with far edit distances. This implies previous works based on masked reconstruction~\cite{haliassos2023raven, kim2023lowresource, shi2022avhubert, Zhu2022VATLMVP} might be imperfect for aligning visual and audio modalities. In contrast, our audio reconstruction objective can be amplified up to ten times above the original task objective, which assists in discerning fine-grained visemes without causing any obstruction. Types of crossmodal indicators, whether they be quantized acoustic tokens from vq-wav2vec~\cite{vqwav2vec} or character-level transcriptions from wav2vec2~\cite{wav2vec2}, are trivial, as seen in Table \ref{tab:word_level}.

\noindent\textbf{Enhancing Speech Representation Learning.} For sentence-level VSR, joint CTC-Attention loss is widely used in previous works~\cite{Afouras2019ASRIA, Ma2021EndConformer, ma2023autoavsr, kim2023lowresource, ma2022visual, haliassos2023raven}. Despite its benefit in the decoding stage, its utility in terms of representation learning remains uncertain. According to Table \ref{tab:ablation}, CTC loss marginally contributes to the perplexity of the model, whereas our audio reconstruction loss term strongly improves learning in both pointcloud and video modalities. The effect of frame-level crossmodal supervision is illustrated in Figure \ref{fig:mean_attention_distance}, where inner representation indicates that the temporal encoder model exhibits a change of bias towards local neighboring frames.

\section{Conclusion}
\label{sec:conclusion}

We addressed the problem of homophenes with an improved crossmodal synchronization method, effectively bridging the divide between visual cues and their corresponding audio segments. The use of quantized audio tokens for direct frame-level supervision enables SyncVSR to achieve state-of-the-art performance on various benchmarks with a remarkable level of data efficiency. We believe SyncVSR is a step toward future developments in the field of multimodal speech recognition.

\section{Acknowledgements}
This work was supported by Institute for Information \& communications Technology Promotion (IITP) grant funded by the Korea government (MSIT) (No.RS-2019-II190075 Artificial Intelligence Graduate School Program (KAIST), Cloud TPUs from Google's TPU Research Cloud (TRC), and an Electronics and Telecommunications Research Institute (ETRI) grant funded by the Korean Government (24ZB1100, Core Technology Research for Self-improving Integrated AI Systems).

\newpage
\bibliographystyle{IEEEtran}
\bibliography{mybib}

\end{document}